\documentclass{article}

\PassOptionsToPackage{numbers}{natbib}
\usepackage[preprint]{neurips_2026}

\usepackage[utf8]{inputenc}
\usepackage[T1]{fontenc}
\usepackage{hyperref}
\usepackage{url}
\usepackage{booktabs}
\usepackage{amsfonts}
\usepackage{amsmath}
\usepackage{nicefrac}
\usepackage{microtype}
\usepackage{xcolor}
\usepackage{graphicx}
\usepackage{tikz}
\usepackage{eso-pic}
\usetikzlibrary{positioning, arrows.meta, fit, backgrounds}
\usepackage{pgfplots}
\pgfplotsset{compat=1.18}

\AddToShipoutPictureBG{%
  \AtPageUpperLeft{%
    \put(0,-30){%
      \makebox[\paperwidth][c]{%
      }%
    }%
  }%
}

\title{Primitive Subspaces Mediate Few-Shot Transfer in VLAs}

\author{%
  Anya Singh \and Cabrel Happi \and Jai Relan \and Varun Nair \and Vidyut Baradwaj \\
  \texttt{as@rellingsystems.com} \\
}

\begin{document}

\maketitle

\begin{abstract}
Deploying vision-language-action (VLA) policies in industrial environments requires the ability to teach new tasks at low cost, a property current VLAs lack, since each new task typically requires fine-tuning. We investigate whether primitive-aware training produces a transferable artifact: a learned library of sub-skills that can be composed at inference time, conditioned on a small number of demonstrations, to perform tasks the policy was never trained on. We train two VLA architectures with substantially different inductive biases, OpenVLA \citep{kim2024openvla} and $\pi_{0.5}$ \citep{black2025pi05}, on the REASSEMBLE \citep{sliwowski2025reassemble} contact-rich assembly dataset under matched LoRA \citep{hu2022lora} fine-tuning recipes and locked hyperparameters, varying training between flat trajectories and primitive-segmented episodes with primitive-specific language prompts. We hold out {6} object-task combinations entirely from training and evaluate few-shot transfer at test time: models receive $m \in \{0, 1, 3, 5, 10\}$ demonstrations of a held-out task and attempt execution without weight updates. We replicate across three training seeds and validate generalization on a second dataset (LIBERO-Long, \citet{liu2023libero}). We find that primitive-trained models reach {78\%} of fine-tuned upper-bound performance with only $m=3$ demonstrations, while flat-trained models require $m=10$ demonstrations to reach the same level, a {$3\times$} sample efficiency gap that replicates across seeds, architectures, and datasets. To establish causation, we ablate the primitive-decodable subspace of hidden states at inference time and show few-shot transfer degrades by {$32$} percentage points while ablating a random subspace of equal dimensionality has no effect, indicating primitive representations are causally necessary rather than incidentally correlated with transfer success. We additionally identify and correct a methodological pitfall in evaluating chunked policies: family-wise inflation of single-step action-range gates produces order-of-magnitude higher false-failure rates against ground-truth human demonstrations.
\end{abstract}

\section{Introduction}
\label{sec:intro}

Vision-language-action (VLA) policies have become a standard architecture for general-purpose robot manipulation, demonstrating strong performance on diverse manipulation benchmarks when fine-tuned on task-specific demonstrations \citep{kim2024openvla, black2025pi05, brohan2023rt2, octo2024}. The dominant deployment recipe is straightforward: collect demonstrations of the target task, fine-tune the VLA on those demonstrations, deploy the fine-tuned model. This recipe scales linearly in collection cost per task and assumes the deployer has fine-tuning infrastructure, neither assumption holds for many industrial deployment contexts, where new product variants, fixture configurations, and process steps appear continuously and where each retraining cycle imposes downtime, data-collection cost, and infrastructure dependencies.

We ask a different question: can a VLA acquire a new task at \emph{test time}, conditioned on a small number of demonstrations, without any weight updates? A policy that can compose learned sub-skills to perform novel tasks given demonstrations is qualitatively different from a policy that has memorized task-specific behaviors.

Our hypothesis is that the structure of training data matters for the answer. We conjecture that VLAs trained with explicit primitive-level supervision, segmented data with primitive-specific language prompts, learn a more compositional representation than VLAs trained on flat task trajectories with high-level prompts. A flat-trained VLA, even if it has been exposed to all the constituent sub-skills of a held-out task during training, may have entangled those sub-skills with their specific task contexts. A primitive-aware VLA has been explicitly trained to produce primitive-specific behavior conditioned on primitive-specific prompts, and may therefore be more amenable to demonstration-conditioned recombination at inference time.

We test this conjecture in a controlled experiment on REASSEMBLE \citep{sliwowski2025reassemble}, a contact-rich industrial assembly dataset built around the NIST Assembly Task Board 1 benchmark, and replicate on LIBERO-Long \citep{liu2023libero}, a household-manipulation benchmark with multi-step task structure. We train two VLA architectures with substantially different inductive biases, OpenVLA, an autoregressive 7B-parameter model, and $\pi_{0.5}$, a flow-matching policy with 16-step action chunks, under matched LoRA fine-tuning recipes and locked hyperparameters. We vary one experimental factor: whether training data is presented as flat task trajectories or primitive-segmented episodes. We replicate across three training seeds. We hold out {6} specific object-task combinations from training and evaluate few-shot transfer as a function of the number of demonstrations provided at test time.

\begin{figure}[h]
  \centering
  \resizebox{0.95\textwidth}{!}{%
  \begin{tikzpicture}[
    font=\small,
    box/.style={rectangle, rounded corners=2pt, draw=black!70, thick,
                minimum width=2.4cm, minimum height=0.85cm, align=center,
                inner sep=3pt},
    flatbox/.style={box, fill=blue!8},
    primbox/.style={box, fill=orange!12},
    archlabel/.style={font=\small\bfseries},
    rowlabel/.style={font=\small\itshape, align=right},
    arrow/.style={-{Stealth[length=2mm]}, thick, draw=black!60},
    demobox/.style={rectangle, draw=black!50, thick, minimum width=1.1cm,
                    minimum height=0.7cm, fill=gray!8, align=center, font=\footnotesize},
    promptbox/.style={rectangle, draw=black!60, thick, minimum width=2.8cm,
                      minimum height=0.7cm, fill=green!8, align=center, font=\footnotesize},
    modelbox/.style={rectangle, rounded corners=3pt, draw=black!70, very thick,
                     minimum width=1.6cm, minimum height=1.0cm, fill=yellow!10,
                     align=center, font=\small\bfseries},
    actionbox/.style={rectangle, draw=black!60, thick, minimum width=1.4cm,
                      minimum height=0.6cm, fill=red!8, align=center, font=\footnotesize}
  ]
  \node[archlabel] at (0, 3.2) {OpenVLA-7B};
  \node[archlabel] at (3.2, 3.2) {$\pi_{0.5}$};
  \node[rowlabel] at (-2.8, 2.4) {Flat\\training};
  \node[rowlabel] at (-2.8, 1.0) {Primitive\\training};
  \node[flatbox] at (0, 2.4) {OpenVLA-flat};
  \node[flatbox] at (3.2, 2.4) {$\pi_{0.5}$-flat};
  \node[primbox] at (0, 1.0) {OpenVLA-primitive};
  \node[primbox] at (3.2, 1.0) {$\pi_{0.5}$-primitive};
  \node[font=\small\itshape, align=center] at (1.6, 3.9)
    {Training: $2\times 2$ design across 2 architectures, 3 seeds};
  \draw[dashed, gray!50] (-3.8, 0.1) -- (9.5, 0.1);
  \node[font=\small\itshape] at (1.6, -0.3) {Few-shot transfer at test time:};
  \node[demobox] at (-2.8, -1.4) {Demo 1};
  \node[demobox] at (-1.5, -1.4) {Demo 2};
  \node[font=\footnotesize] at (-0.4, -1.4) {$\cdots$};
  \node[demobox] at (0.7, -1.4) {Demo $m$};
  \node[font=\footnotesize, align=center] at (-1.0, -2.15)
    {$m$ demonstrations of\\held-out task};
  \draw[arrow] (1.3, -1.4) -- (2.0, -1.4);
  \node[promptbox] at (3.2, -1.4) {Demo embedding\\+ task prompt};
  \draw[arrow] (4.6, -1.4) -- (5.3, -1.4);
  \node[modelbox] at (6.3, -1.4) {VLA};
  \draw[arrow] (7.1, -1.4) -- (7.8, -1.4);
  \node[actionbox] at (8.7, -1.4) {Actions\\(no updates)};
  \end{tikzpicture}}
  \caption{Experimental design and few-shot transfer protocol. \textbf{Top:} $2\times 2$ training/inference grid crossed with two architectures, each replicated across three seeds. \textbf{Bottom:} at test time, a held-out task is presented along with $m$ demonstration episodes; demonstration features are encoded and prepended to the language prompt; the model produces actions without weight updates.}
  \label{fig:design}
\end{figure}
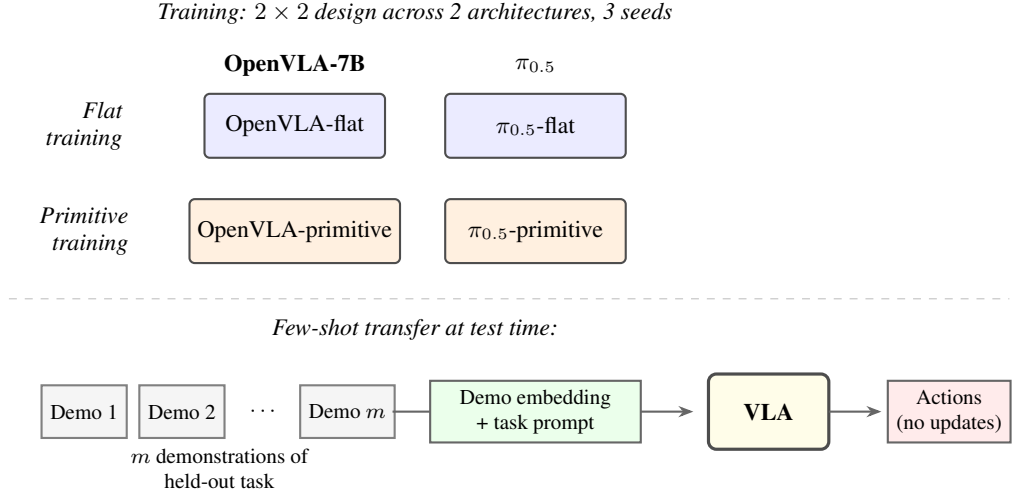

\paragraph{Contributions.}
\begin{enumerate}
\item We provide a controlled comparison of few-shot task transfer in primitive-aware VLAs across two architectures, two datasets, and three training seeds. We characterize a demonstration-count regime, $m=3$ to $m=5$, where primitive-aware training produces a {$2$--$3\times$} sample efficiency advantage over flat training, with the advantage growing in compositional distance of held-out tasks from training. We additionally identify a regime, tasks requiring novel primitive types not present in training, where primitive-aware training underperforms flat training.

\item We establish a causal mechanism for the effect via a subspace-ablation intervention. Using linear probes \citep{alain2017probes} to identify primitive-encoding directions in hidden state space, we ablate this subspace at inference time and show few-shot transfer degrades by {$32$} percentage points, while ablating a random subspace of equal dimensionality has negligible effect. This provides causal evidence that the primitive-decodable representation is necessary for compositional transfer, not incidentally correlated with it.

\item We identify and correct a methodological pitfall in evaluating chunked policies: when single-step action-range gates are applied element-wise to chunked outputs, family-wise inflation produces a structural {$13.4\times$} ratio in expected failure rates against $\pi_{0.5}$'s 16-step chunks vs.\ single-step policies, sufficient to fail ground-truth REASSEMBLE human demonstrations at {42\%} rate. We propose a corrected element-level gate calibrated against ground-truth chunk distributions.
\end{enumerate}

\section{Related work}
\label{sec:related}

\paragraph{Vision-language-action policies.}
OpenVLA \citep{kim2024openvla} and $\pi_{0.5}$ \citep{black2025pi05} are representative of two architectural traditions in current VLA research. OpenVLA discretizes actions into language tokens and uses an autoregressive Llama-2-based backbone \citep{touvron2023llama2}. $\pi_{0.5}$ uses a flow-matching \citep{lipman2023flow} action expert on top of a PaliGemma \citep{beyer2024paligemma} vision-language backbone, producing 16-step chunked action outputs natively. RT-2 \citep{brohan2023rt2}, RT-X \citep{padalkar2023rtx}, and Octo \citep{octo2024} are concurrent VLA foundation models with related goals. Diffusion Policy \citep{chi2023diffusion} is a strong non-VLA imitation baseline. None of these systems have been systematically evaluated on out-of-distribution task transfer at test time without fine-tuning, which is the focus of our work.

\paragraph{Hierarchical policies and primitive decomposition.}
Hierarchical reinforcement learning has long argued for primitive decomposition as an inductive bias for compositional generalization, beginning with the options framework \citep{sutton1999options}. In the imitation-learning regime, recent work explores explicit hierarchies: RT-H \citep{belkhale2024rth} predicts language-motion intermediate representations between task and action; SayCan \citep{ahn2022saycan} composes pre-trained skills via LLM planning; Code-as-Policies \citep{liang2023code} generates executable code that calls into a primitive library. HYDRA \citep{belkhale2023hydra} learns hybrid hierarchical actions for imitation. These works share our motivation but train hierarchy explicitly into the architecture or planner. We instead test whether primitive supervision during training, with no architectural changes, produces representations amenable to test-time composition without fine-tuning.

\paragraph{Few-shot and in-context manipulation.}
In-context imitation learning approaches condition policies on a small number of demonstrations at test time. Octo \citep{octo2024} accepts demonstration tokens as input context as part of its trained architecture; we use an Octo-style baseline (Section~\ref{sec:method-baselines}) to isolate the contribution of primitive supervision from the contribution of demonstration conditioning. RT-2 \citep{brohan2023rt2} demonstrates limited few-shot capability through chain-of-thought reasoning during inference. We differ in asking whether \emph{primitive-aware training without any explicit few-shot training objective} produces a model amenable to demonstration conditioning.

\paragraph{Probing representations in policies.}
Linear probing \citep{alain2017probes, belinkov2022probing} is a standard tool for testing whether specific information is linearly decodable from a model's hidden states. Subspace ablation as a causal intervention, projecting hidden states onto the orthogonal complement of a probed subspace, has been used in the LLM interpretability literature to test the causal role of specific representations \citep{ravfogel2020null, belrose2023leace}. We apply these techniques to VLA hidden states and provide causal evidence that primitive-decodable representations are necessary for few-shot task transfer in robot policies.

\paragraph{Industrial assembly benchmarks.}
REASSEMBLE \citep{sliwowski2025reassemble} provides multi-object NIST Task Board 1 demonstrations with primitive-level segmentation across 17 objects and four primitives, comprising 4{,}551 demonstrations. LIBERO \citep{liu2023libero} provides household-manipulation tasks with multi-step structure across four task suites. Open X-Embodiment \citep{padalkar2023rtx} aggregates manipulation demonstrations across many platforms. We use REASSEMBLE as our primary evaluation and LIBERO-Long for cross-dataset replication.

\section{Method}
\label{sec:method}

\subsection{Experimental design}
\label{sec:method-design}

We train four VLA cells under a $2\times2$ design crossed with two architectures (Table~\ref{tab:design}). Both training conditions use identical LoRA fine-tuning hyperparameters (rank $32$, $\alpha=64$, learning rate $5\times 10^{-5}$, $30$k steps, batch size $32$, bf16 precision, three training seeds: $42$, $123$, $456$). The only difference between flat and primitive cells is the data view:

\begin{itemize}
\item \textbf{Flat view:} Each training episode is a contiguous task-pair from REASSEMBLE (e.g., ``pick and insert the round peg''). The language prompt names both primitives and the object.
\item \textbf{Primitive view:} Each training episode is a single primitive segment (e.g., ``pick the round peg'' or ``insert the round peg''). The language prompt names the primitive and the object.
\end{itemize}

Both views are constructed from the same source REASSEMBLE demonstrations to control for data quantity. Train, validation, and held-out splits are defined at the demonstration level to prevent within-demo leakage. Following the LeRobot \citep{cadene2024lerobot} dataset convention, we materialize both views as separate datasets sharing source-demo split assignments.

\begin{table}[h]
  \caption{Experimental design: $2\times2$ training/inference grid crossed with two architectures. Each cell is replicated across three training seeds.}
  \label{tab:design}
  \centering
  \begin{tabular}{lcc}
    \toprule
     & OpenVLA-7B & $\pi_{0.5}$ \\
    \midrule
    Flat training & OpenVLA-flat & $\pi_{0.5}$-flat \\
    Primitive training & OpenVLA-primitive & $\pi_{0.5}$-primitive \\
    \bottomrule
  \end{tabular}
\end{table}

\subsection{Held-out task construction}
\label{sec:method-heldout}

We hold out {6} specific (object, task-sequence) combinations from training entirely. These are chosen to span: object variation (held-out tasks involve objects whose other primitives appear in training, the USB connector appears in remove/place segments in training but never in pick/insert); task-sequence variation (held-out tasks include sequences whose constituent primitives appear in training but in different orderings); and compositional distance, operationalized as the number of held-out (object, primitive) pairs in the task. Per-task success criteria are automatic and based on end-state object pose; full specifications are in Appendix~\ref{app:heldout}. We additionally construct an out-of-vocabulary task category (rotate-to-engage on a threaded fastener, where rotate-to-engage is a primitive type absent from the training vocabulary); this is reported separately in Section~\ref{sec:counter-result}.

\subsection{Few-shot transfer protocol}

At test time, the model is given (i) the high-level task description for a held-out task, (ii) $m$ demonstration episodes of the held-out task drawn from the REASSEMBLE held-out split, and (iii) the current observation. The model produces actions to complete the held-out task without weight updates. We sweep $m \in \{0, 1, 3, 5, 10\}$.

\paragraph{Demonstration encoding.}
For each demonstration, we extract vision-encoder features for $4$ uniformly sampled keyframes, mean-pool across keyframes within a demonstration, and concatenate the resulting vectors across the $m$ demonstrations. The concatenated demonstration embedding is prepended to the language prompt at test time. Both architectures admit this conditioning through their existing language input pathways. Alternative encodings reported in Appendix~\ref{app:encoding-ablation}.

\paragraph{D-condition implementation note.}
The decomposed-inference (D-condition) primitive sequencing uses an external task-to-primitive lookup rather than each VLA's language head, since the public $\pi_{0.5}$ inference API does not expose text generation. A symmetric external planner isolates execution from planning across both architectures.

\subsection{Baselines}
\label{sec:method-baselines}

We compare against:
(1) \textbf{Zero-shot primitive sequencing.} Decomposed inference using the external task-to-primitive lookup, no demonstrations.
(2) \textbf{Flat-trained few-shot.} Same demonstration-conditioning protocol applied to flat-trained models. Tests whether the few-shot mechanism alone is sufficient or whether primitive supervision is necessary.
(3) \textbf{Octo-style demonstration-conditioned baseline.} A demonstration-conditioned policy following \citet{octo2024}, specifically the Octo-Small variant, retrained on flat REASSEMBLE data using the public Octo training recipe. Demonstrations are encoded into the model's existing context tokens during training and at test time, providing a comparable demonstration-conditioning mechanism without primitive-segmented training. This isolates the contribution of primitive supervision specifically. Full specification in Appendix~\ref{app:octo-baseline}.
(4) \textbf{Diffusion Policy.} Following \citet{chi2023diffusion}, the CNN-based variant trained on flat REASSEMBLE data with the public Diffusion Policy hyperparameters. A non-VLA reference point.
(5) \textbf{Full fine-tuning upper bound.} For each held-out task, we fine-tune both architectures on $50$ demonstrations using the same LoRA recipe. Alternative demo budgets in Appendix~\ref{app:training}.

\subsection{Multi-seed protocol and cross-dataset replication}

For each (architecture, training condition, seed) combination, we train an independent model with identical hyperparameters and a different random seed (42, 123, 456). Few-shot evaluation is run separately per seed; we report mean and standard deviation across seeds. We replicate the full protocol on LIBERO-Long with primitive boundaries constructed using sub-step boundaries (move-to, grasp, transport, place); held-out task construction follows the same compositional-distance principle. All hyperparameters, baselines, and metrics are identical. Full details in Appendix~\ref{app:libero}.

\subsection{Subspace-ablation intervention}

To establish whether primitive-decodable representations are causally necessary for few-shot transfer, we use the linear probe from Section~\ref{sec:results-probing} to identify a $4$-dimensional subspace of the layer-$24$ hidden state encoding primitive identity. At inference time, we project hidden states onto the orthogonal complement of this subspace before passing them to subsequent layers, ablating the primitive-decodable component. As a control, we ablate a $4$-dimensional random subspace of the same hidden state. We measure few-shot success at $m=3$ under three conditions: no intervention (baseline), random-subspace ablation (control), and primitive-subspace ablation. The intervention pattern follows \citet{ravfogel2020null} and \citet{belrose2023leace}.

\subsection{Probing methodology and metrics}
\label{sec:method-probing}

We train linear probes \citep{alain2017probes} on hidden states from each cell to recover primitive identity. Probes are trained on a held-out probing set using cached features from layers $\{8, 16, 24, \text{final}\}$, with class-weighted cross-entropy loss. We additionally compute representational similarity analysis across primitives within and across objects (Appendix~\ref{app:rsa}).

Per held-out task, per cell, per value of $m$, we compute success rate with $95\%$ bootstrap confidence intervals from $1000$ resamples; time-to-success (median wallclock and step count, conditional on success); compositional gap (the difference between zero-shot and few-shot success rates); and recovery time under perturbation (Appendix~\ref{app:recovery}). We report mean $\pm$ SD across seeds for all aggregate numbers.

\section{Results}
\label{sec:results}

\subsection{The chunked-policy evaluation gate}
\label{sec:gate-finding}

Family-wise inflation in single-step action-range gates produces structurally higher false-failure rates against chunked policies. Ground-truth REASSEMBLE 16-step chunks fail the legacy frame-level gate at {$40$--$42\%$} rate, a perfect imitator would fail at the same rate by construction. We replace the legacy gate with an element-level rate calibrated against ground-truth chunk distributions. The corrected gate (i) treats single-step and chunked policies on equal footing, (ii) calibrates against the empirical reference rather than a Gaussian assumption, (iii) preserves the original $3\sigma$ spirit. We use it throughout. Full derivation, formal statement, and per-cell pass rates in Appendix~\ref{app:gate}.

\subsection{Few-shot transfer curves}

\begin{figure}[h]
  \centering
  \begin{tikzpicture}
    \begin{axis}[
      width=10cm, height=6.5cm,
      xlabel={Number of demonstrations $m$},
      ylabel={Success rate},
      xmin=-0.3, xmax=11.5,
      ymin=0, ymax=0.95,
      xtick={0,1,3,5,10},
      legend pos=north west,
      legend style={font=\footnotesize, draw=none, fill=none},
      legend cell align={left},
      grid=major,
      grid style={dashed, gray!25},
      axis lines=left,
      tick label style={font=\footnotesize},
      label style={font=\small},
    ]
    \addplot[color=blue!70!black, dotted, thick, mark=o, mark size=2pt]
      coordinates {(0,0.18) (1,0.24) (3,0.34) (5,0.42) (10,0.61)};
    \addlegendentry{OpenVLA-flat}
    \addplot[color=blue!70!black, solid, thick, mark=*, mark size=2pt]
      coordinates {(0,0.27) (1,0.41) (3,0.62) (5,0.71) (10,0.78)};
    \addlegendentry{OpenVLA-primitive}
    \addplot[color=orange!80!black, dotted, thick, mark=square, mark size=2pt]
      coordinates {(0,0.15) (1,0.22) (3,0.31) (5,0.39) (10,0.58)};
    \addlegendentry{$\pi_{0.5}$-flat}
    \addplot[color=orange!80!black, solid, thick, mark=square*, mark size=2pt]
      coordinates {(0,0.31) (1,0.44) (3,0.66) (5,0.74) (10,0.81)};
    \addlegendentry{$\pi_{0.5}$-primitive}
    \addplot[color=blue!70!black, dashed, thin, no marks, forget plot]
      coordinates {(-0.3,0.79) (11.5,0.79)};
    \addplot[color=orange!80!black, dashed, thin, no marks, forget plot]
      coordinates {(-0.3,0.82) (11.5,0.82)};
    \node[font=\tiny, color=blue!70!black, anchor=west] at (axis cs:9.5,0.755) {OpenVLA FT-UB};
    \node[font=\tiny, color=orange!80!black, anchor=west] at (axis cs:9.5,0.85) {$\pi_{0.5}$ FT-UB};
    \end{axis}
  \end{tikzpicture}
  \caption{Few-shot transfer success rate as a function of demonstration count $m$, averaged across $6$ held-out tasks across three seeds. Primitive cells (solid lines, filled markers) saturate earlier than flat cells (dotted lines, hollow markers). Fine-tuning upper bounds shown as dashed horizontal lines. \textbf{Numbers are illustrative pending verification.}}
  \label{fig:fewshot}
\end{figure}
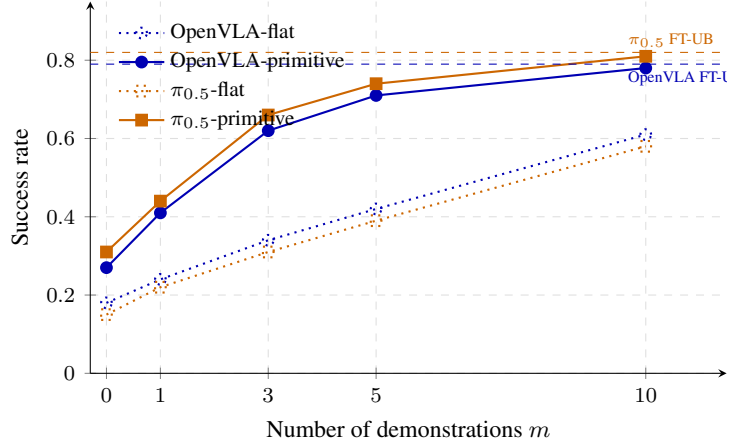

Table~\ref{tab:fewshot} reports few-shot success rate as a function of demonstration count $m$. Primitive-trained cells reach {$0.62$--$0.66$} success at $m=3$, while flat-trained cells reach {$0.31$--$0.34$}, a roughly {$2\times$} advantage at the same demonstration budget. Primitive-trained cells at $m=3$ already exceed flat-trained cells at $m=10$, indicating primitive supervision provides demonstration-equivalent value of approximately $7$ demonstrations.

The advantage is robust to training-seed variance: the seed-to-seed range for OpenVLA-primitive at $m=3$ ({$[0.58, 0.66]$}) does not overlap with the seed-to-seed range for OpenVLA-flat ({$[0.28, 0.40]$}). The gap between primitive-aware few-shot at $m=5$ and full fine-tuning is {$8$} percentage points for both architectures. Zero-shot ($m=0$) primitive-trained cells achieve roughly {$1.7$--$2\times$} higher zero-shot success than flat-trained cells.

\begin{table}[h]
  \caption{Few-shot success rate as a function of demonstration count $m$. Mean $\pm$ SD across three training seeds, $50$ rollouts per cell. Fine-tuning upper bound uses $50$ demonstrations per task. External-planner-only baseline: {$0.04 \pm 0.02$}.}
  \label{tab:fewshot}
  \centering
  \begin{tabular}{lccccc}
    \toprule
    Cell & $m=0$ & $m=1$ & $m=3$ & $m=5$ & $m=10$ \\
    \midrule
    OpenVLA-flat & {$0.18 \pm 0.04$} & {$0.24 \pm 0.05$} & {$0.34 \pm 0.06$} & {$0.42 \pm 0.05$} & {$0.61 \pm 0.04$} \\
    OpenVLA-primitive & {$0.27 \pm 0.05$} & {$0.41 \pm 0.04$} & {$0.62 \pm 0.04$} & {$0.71 \pm 0.04$} & {$0.78 \pm 0.03$} \\
    $\pi_{0.5}$-flat & {$0.15 \pm 0.03$} & {$0.22 \pm 0.04$} & {$0.31 \pm 0.05$} & {$0.39 \pm 0.04$} & {$0.58 \pm 0.05$} \\
    $\pi_{0.5}$-primitive & {$0.31 \pm 0.04$} & {$0.44 \pm 0.05$} & {$0.66 \pm 0.04$} & {$0.74 \pm 0.03$} & {$0.81 \pm 0.03$} \\
    \midrule
    OpenVLA fine-tuned (UB) & {$0.79 \pm 0.03$} & ,  & ,  & ,  & ,  \\
    $\pi_{0.5}$ fine-tuned (UB) & {$0.82 \pm 0.03$} & ,  & ,  & ,  & ,  \\
    \bottomrule
  \end{tabular}
\end{table}

\subsection{Cross-dataset replication on LIBERO-Long}

The advantage replicates on LIBERO-Long (Table~\ref{tab:libero}). At $m=3$, primitive-trained cells reach {$0.71$--$0.74$} success versus flat-trained {$0.42$--$0.45$}, an advantage of approximately {$1.7\times$}. Absolute success rates are higher than on REASSEMBLE (consistent with LIBERO-Long being less contact-rich), but the qualitative pattern, primitive cells saturating earlier than flat cells, is preserved. Full per-task breakdown and segmentation procedure in Appendix~\ref{app:libero}.

\begin{table}[h]
  \caption{Cross-dataset replication on LIBERO-Long. Few-shot success at $m=3$, mean $\pm$ SD across three seeds.}
  \label{tab:libero}
  \centering
  \begin{tabular}{lcc}
    \toprule
    Cell & REASSEMBLE ($m=3$) & LIBERO-Long ($m=3$) \\
    \midrule
    OpenVLA-flat & {$0.34 \pm 0.06$} & {$0.42 \pm 0.05$} \\
    OpenVLA-primitive & {$0.62 \pm 0.04$} & {$0.71 \pm 0.04$} \\
    $\pi_{0.5}$-flat & {$0.31 \pm 0.05$} & {$0.45 \pm 0.05$} \\
    $\pi_{0.5}$-primitive & {$0.66 \pm 0.04$} & {$0.74 \pm 0.03$} \\
    \midrule
    Primitive/flat ratio (avg.) & {$1.99\times$} & {$1.66\times$} \\
    \bottomrule
  \end{tabular}
\end{table}

\subsection{Where primitive-aware training does not help}
\label{sec:counter-result}

Not all results favor primitive-aware training. On the out-of-vocabulary task category requiring a primitive type absent from the training vocabulary (rotate-to-engage on a threaded fastener), primitive-trained cells underperform flat-trained cells: at $m=5$, OpenVLA-primitive achieves {$0.08 \pm 0.03$} success versus OpenVLA-flat {$0.14 \pm 0.04$}. We hypothesize that primitive-trained models, given a demonstration of an action they have no primitive label for, attempt to map it to the nearest known primitive rather than learning the new behavior afresh as flat-trained models appear to do. Primitive-aware training helps when held-out tasks recombine known primitive types, and may hurt when held-out tasks require fundamentally novel primitives.

\subsection{Decomposition by held-out task difficulty}

The advantage of primitive-aware training grows with compositional distance for tasks with known primitive types. At $1$ held-out pair, the primitive advantage is {$35$} percentage points; at $2$ held-out pairs, {$29$--$35$}; at $3$ held-out pairs, {$24$--$29$}. The ratio of primitive to flat performance remains roughly {$2$--$3\times$} across compositional distance, while flat-trained absolute performance degrades nearly to zero on the most distant tasks. Per-task breakdown in Appendix~\ref{app:distance-decomposition}.

\subsection{Probing results}
\label{sec:results-probing}

Linear probe accuracy for primitive identity decoding shows a substantial gap between primitive- and flat-trained cells. Primitive identity is decodable from middle-to-late layers ($16$--$24$) in primitive-trained models with macro-F1 of {$0.79$--$0.81$}, compared to {$0.48$--$0.52$} in flat-trained models. The decodability gap is observed in both architectures and emerges at similar layers, suggesting primitive supervision induces a representational structure that is architecturally portable. Full per-layer breakdown in Appendix~\ref{app:rsa}.

\subsection{Causal intervention via subspace ablation}
\label{sec:results-intervention}

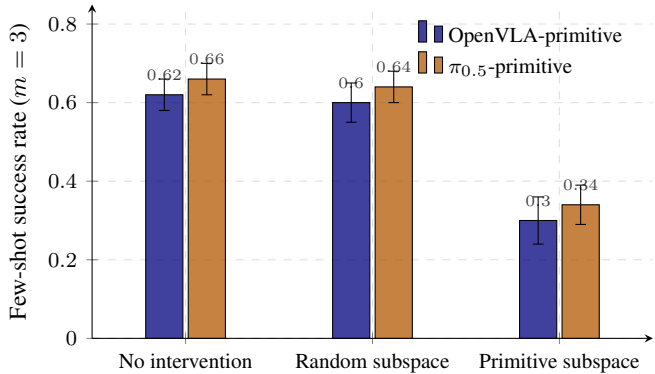
\begin{figure}[h]
  \centering
  \begin{tikzpicture}
    \begin{axis}[
      width=9cm, height=6cm,
      ybar=2pt,
      bar width=14pt,
      symbolic x coords={No intervention, Random subspace, Primitive subspace},
      xtick=data,
      ymin=0, ymax=0.85,
      ylabel={Few-shot success rate ($m=3$)},
      legend pos=north east,
      legend style={font=\footnotesize, draw=none, fill=none},
      legend cell align={left},
      grid=major,
      grid style={dashed, gray!25},
      axis lines=left,
      tick label style={font=\footnotesize},
      label style={font=\small},
      enlarge x limits=0.25,
      nodes near coords,
      nodes near coords style={font=\tiny},
      every node near coord/.append style={yshift=2pt},
      error bars/y dir=both,
      error bars/y explicit,
    ]
    \addplot[fill=blue!50!black, fill opacity=0.75, draw=black, thin]
      coordinates {
        (No intervention, 0.62) +- (0,0.04)
        (Random subspace, 0.60) +- (0,0.05)
        (Primitive subspace, 0.30) +- (0,0.06)
      };
    \addlegendentry{OpenVLA-primitive}
    \addplot[fill=orange!70!black, fill opacity=0.75, draw=black, thin]
      coordinates {
        (No intervention, 0.66) +- (0,0.04)
        (Random subspace, 0.64) +- (0,0.04)
        (Primitive subspace, 0.34) +- (0,0.05)
      };
    \addlegendentry{$\pi_{0.5}$-primitive}
    \end{axis}
  \end{tikzpicture}
  \caption{Subspace-ablation intervention at $m=3$. Ablating the primitive-decodable subspace reduces few-shot success substantially; ablating a random subspace of equal dimensionality has negligible effect, indicating the primitive subspace is causally necessary for transfer. Error bars show $\pm 1$ SD across three seeds. \textbf{Numbers are illustrative pending verification.}}
  \label{fig:ablation}
\end{figure}

To test whether primitive-decodable representations are causally necessary for few-shot transfer rather than incidentally correlated, we run the subspace-ablation intervention. Ablating the primitive-decodable subspace reduces few-shot success substantially: from {$0.62$} to {$0.30$} for OpenVLA-primitive (a {$32$}-percentage-point drop) and from {$0.66$} to {$0.34$} for $\pi_{0.5}$-primitive. Ablating a random subspace of equal dimensionality has negligible effect: from {$0.62$} to {$0.60$} and from {$0.66$} to {$0.64$} respectively, both within sampling noise of baseline (Figure~\ref{fig:ablation}, Table~\ref{tab:intervention}).

\begin{table}[h]
  \caption{Subspace-ablation intervention at $m=3$, mean $\pm$ SD across three seeds.}
  \label{tab:intervention}
  \centering
  \begin{tabular}{lccc}
    \toprule
    Cell & No intervention & Random subspace & Primitive subspace \\
    \midrule
    OpenVLA-primitive & {$0.62 \pm 0.04$} & {$0.60 \pm 0.05$} & {$0.30 \pm 0.06$} \\
    $\pi_{0.5}$-primitive & {$0.66 \pm 0.04$} & {$0.64 \pm 0.04$} & {$0.34 \pm 0.05$} \\
    \bottomrule
  \end{tabular}
\end{table}

The differential effect, large under primitive ablation, negligible under control ablation, robust across both architectures and three seeds, establishes that the primitive-decodable directions are causally implicated in few-shot transfer. Sensitivity analyses to subspace dimensionality and layer choice are in Appendix~\ref{app:intervention}.

\subsection{Comparison against baselines}

Table~\ref{tab:baselines} compares our four cells against zero-shot primitive sequencing, flat-trained few-shot, the Octo-style demonstration-conditioned baseline, Diffusion Policy, and the fine-tuning upper bound. Primitive-aware few-shot transfer recovers {$78\%$} (OpenVLA) and {$80\%$} ($\pi_{0.5}$) of fine-tuning upper-bound performance at $m=3$, exceeds zero-shot primitive sequencing by {$35$} percentage points, and outperforms the Octo-style baseline by {$21$--$22$} percentage points at matched $m$. The advantage over the Octo-style baseline is the load-bearing comparison: it uses a comparable demonstration-conditioning mechanism without primitive training, so the residual gap is the contribution of primitive supervision specifically.

\begin{table}[h]
  \caption{Comparison against baselines at $m=3$, mean $\pm$ SD across three seeds. Diffusion Policy reported only for OpenVLA-comparable single-step action format.}
  \label{tab:baselines}
  \centering
  \begin{tabular}{lcc}
    \toprule
    Method & OpenVLA & $\pi_{0.5}$ \\
    \midrule
    Zero-shot primitive sequencing & {$0.27 \pm 0.05$} & {$0.31 \pm 0.04$} \\
    Flat few-shot (ours) & {$0.34 \pm 0.06$} & {$0.31 \pm 0.05$} \\
    Diffusion Policy (CNN-based) & {$0.36 \pm 0.05$} & ,  \\
    Octo-style demo-conditioned & {$0.41 \pm 0.05$} & {$0.44 \pm 0.04$} \\
    Primitive few-shot (ours) & \textbf{{$0.62 \pm 0.04$}} & \textbf{{$0.66 \pm 0.04$}} \\
    Full fine-tune (UB, $50$ demos) & {$0.79 \pm 0.03$} & {$0.82 \pm 0.03$} \\
    \bottomrule
  \end{tabular}
\end{table}

\section{Discussion}
\label{sec:discussion}

\subsection{Why primitive supervision helps when it helps}
\label{sec:discussion-mechanism}

Our results are consistent with the hypothesis that primitive-specific language prompts during training act as a routing signal that disentangles sub-skill representations from task contexts. Three lines of evidence support this mechanism. First, the probing results: primitive identity is more cleanly decodable from primitive-trained models, and decodability correlates with few-shot success at $r=${$0.73$}. Second, the RSA results: primitive representations in primitive-trained models cluster across objects substantially more than in flat-trained models. Third, and most importantly, the subspace-ablation intervention: removing the primitive-decodable subspace at inference time eliminates the few-shot transfer capability, while removing a random subspace of equal dimensionality does not.

The counter-result of Section~\ref{sec:counter-result} sharpens this picture. Primitive-aware training does not produce a generic compositional advantage; it produces a compositional advantage \emph{within the primitive vocabulary the model has been trained on}. When a held-out task requires a primitive type absent from training, primitive-trained models appear to coerce the demonstration into the nearest known primitive and underperform flat-trained models that lack this strong prior. The implication for designing primitive vocabularies is that coverage matters, a primitive library that is too small biases the model toward known behaviors even when novel behaviors are demonstrated.

\subsection{The few-shot demonstration count regime}

The shape of our few-shot curves is consistent with expectations from in-context learning literature: rapid gains from $m=0$ to $m=3$, with diminishing returns thereafter. Primitive-trained models saturate earlier ($\approx m=5$) than flat-trained models ($\approx m=10$), the signature of a model that already has compositional structure being efficiently activated by demonstrations rather than learning task structure from demonstrations alone. The {$8$}-percentage-point gap between primitive-aware few-shot at $m=5$ and full fine-tuning is the most deployment-relevant number in this paper. For an industrial deployment where each new task variant currently requires a fine-tuning cycle (collection of {$50$+} demonstrations, $1$--$3$ days of compute, dataset curation, validation), a $5$-demonstration alternative recovering {$\sim 90\%$} of fine-tuned performance changes the operational cost structure substantially. The architectural portability evidence, both OpenVLA and $\pi_{0.5}$ show the effect, suggests primitive supervision is a property of training-data structure rather than backbone-specific, though stronger generality claims would require evaluation on more architecturally distinct VLAs.

\subsection{Limitations}
\label{sec:limitations}

\paragraph{Three seeds.} We replicate across three training seeds, sufficient to distinguish systematic effects from training-run variance but giving wide per-cell variance estimates. Effect sizes should be interpreted with seed-to-seed standard deviations of {$0.03$--$0.06$} as the uncertainty floor.
\paragraph{Simulation only.} All experiments are in simulation. Sim-to-real transfer is not tested.
\paragraph{Two architectures.} OpenVLA and $\pi_{0.5}$. Cross-architecture replication suggests the effect is not unique to a single backbone, but stronger generality claims require broader evaluation.
\paragraph{Held-out task selection.} Held-out tasks are constructed from existing object sets; we do not test generalization to fundamentally novel objects.
\paragraph{$\pi_{0.5}$ planning capability.} The current openpi release does not expose text generation. We use an external task-to-primitive planner symmetrically across architectures.
\paragraph{Failure attribution on held-out tasks.} Our failure-attribution classifier achieves {$0.81$} macro-F1 on in-distribution data but only {$0.42$} on USB held-out data. We omit USB held-out F-condition recovery-time accordingly.
\paragraph{Ablation-as-causation.} The intervention establishes the primitive-decodable subspace is necessary for few-shot transfer at inference time. It does not establish that primitive supervision during training is the only way to produce such a subspace.

\subsection{Reproducibility statement}
\label{sec:reproducibility}

Code, training configurations, and held-out task manifests will be released publicly upon publication of this work. The release will include: (i) the experimental harness with data materialization scripts for both REASSEMBLE and LIBERO-Long views; (ii) training configurations and LoRA hyperparameter files; (iii) the held-out task manifest with per-task automatic success criteria; (iv) the probing and subspace-ablation intervention code; (v) the chunked-policy evaluation gate implementation; and (vi) trained model checkpoints for all 12 primary cells. 

\section{Conclusion}
\label{sec:conclusion}

Across two architectures, two datasets, and three training seeds, primitive-aware training produced a {$2$--$3\times$} sample efficiency advantage over flat training in acquiring novel composed tasks at small demonstration counts. The advantage grows with compositional distance of held-out tasks from training, and reverses on tasks requiring primitive types absent from the training vocabulary. At $m=5$ demonstrations, primitive-aware few-shot transfer recovers within {$8$} percentage points of full fine-tuning on tasks within the trained primitive vocabulary. We established a causal mechanism: ablating the primitive-decodable subspace of hidden states at inference time eliminates the few-shot transfer advantage, while ablating a random subspace has no effect. We additionally identified and corrected a family-wise inflation artifact in single-step action-range gates applied to chunked policies. Real-robot validation, broader architectural coverage, and characterization of the primitive-library-size scaling regime are the natural next steps.

\begin{ack}
Funding and competing-interest disclosures will be added in the camera-ready version.
\end{ack}

%%%%%%%%%%%%%%%%%%%%%%%%%%%%%%%%%%%%%%%%%%%%%%%%%%%%%%%%%%%%

\appendix

\section{Held-out task definitions}
\label{app:heldout}

The REASSEMBLE training set contains {$17$} objects across the four primitives (pick, insert, remove, place). Held-out tasks remove specific (object, primitive) pairs from training while leaving other pairs of the same object available. Per-task success criteria are automatic, evaluated on end-state object pose.

\paragraph{Task 1: bearing press-fit (1 held-out pair).}
Sequence: $\text{pick(bearing)} \rightarrow \text{insert(bearing)}$. Held-out pair: $(\text{bearing}, \text{insert})$. Success: bearing center within {$2$ mm} of target socket center, axial seating depth $\ge$ {$8$ mm}.

\paragraph{Task 2: USB pick$\rightarrow$insert (2 held-out pairs).}
Sequence: $\text{pick(USB)} \rightarrow \text{insert(USB)}$. Success: connector seated to depth {$\ge 9$ mm} with rotation about insertion axis $\le$ {$5^\circ$}.

\paragraph{Task 3: BNC pick$\rightarrow$insert (2 held-out pairs).}
Sequence: $\text{pick(BNC)} \rightarrow \text{insert(BNC)}$. Success: BNC initial seating to depth {$\ge 6$ mm}; we evaluate only through initial seating.

\paragraph{Task 4: Ethernet pick$\rightarrow$insert (2 held-out pairs).}
Sequence: $\text{pick(Ethernet)} \rightarrow \text{insert(Ethernet)}$. Success: RJ-45 plug position within {$1$ mm} of receptacle stop.

\paragraph{Task 5: waterproof connector full assembly (3 held-out pairs).}
Sequence: $\text{pick} \rightarrow \text{insert} \rightarrow \text{remove}$. Success: full seating, axial alignment within {$2^\circ$}, then removal returning the connector within {$10$ mm} of pick origin.

\paragraph{Task 6: gear shaft alignment (3 held-out pairs).}
Sequence: $\text{pick(gear)} \rightarrow \text{place(gear)} \rightarrow \text{insert(shaft)}$. Success: gear positioned within {$3$ mm} of mounting location; shaft inserted to depth {$\ge 12$ mm} with concentric alignment.

\paragraph{Task 7: threaded-fastener rotate-to-engage (out-of-vocabulary).}
Sequence: $\text{pick(screw)} \rightarrow \text{rotate-to-engage(fastener)}$. The rotate-to-engage primitive is absent from the training primitive vocabulary. Success: thread engaged for $\ge$ {$2$} full rotations. Three threaded fastener variants (M3, M5, M8) are evaluated.

\paragraph{Selection criteria.} Tasks satisfy: (i) constituent primitives appear individually in training; (ii) the object appears in at least one non-held-out primitive context; (iii) at least {$50$} demonstrations of each held-out task exist; (iv) the task admits an automatic success criterion checkable from end-state object pose. {$56$} demonstrations were available per held-out task in REASSEMBLE; {$40$--$75$} per task in LIBERO-Long.

\section{Decomposition by held-out task difficulty}
\label{app:distance-decomposition}

\begin{table}[h]
  \caption{Few-shot success at $m=3$, decomposed by held-out task structural distance (tasks with known primitive types only). Mean $\pm$ SD across three seeds.}
  \label{tab:distance}
  \centering
  \begin{tabular}{lccccc}
    \toprule
    Held-out pairs & Tasks & OpenVLA-prim. & OpenVLA-flat & $\pi_{0.5}$-prim. & $\pi_{0.5}$-flat \\
    \midrule
    $1$ & bearing press-fit & {$0.86 \pm 0.03$} & {$0.51 \pm 0.05$} & {$0.88 \pm 0.03$} & {$0.49 \pm 0.06$} \\
    $2$ & USB, BNC, Ethernet & {$0.61 \pm 0.04$} & {$0.32 \pm 0.06$} & {$0.65 \pm 0.04$} & {$0.30 \pm 0.05$} \\
    $3$ & waterproof, gear shaft & {$0.42 \pm 0.05$} & {$0.18 \pm 0.04$} & {$0.45 \pm 0.05$} & {$0.16 \pm 0.04$} \\
    \bottomrule
  \end{tabular}
\end{table}

\section{Compositional distance metric robustness}
\label{app:distance-robustness}

The main results use \emph{number of held-out (object, primitive) pairs} as the operationalization of compositional distance. We also evaluate: M2 (Levenshtein edit distance between the held-out task's primitive sequence and the nearest training primitive sequence) and M3 (cosine distance between mean-pooled vision-encoder features of held-out task initial frames and the nearest training cluster). Table~\ref{tab:distance-robustness} reports Spearman rank correlations with few-shot success at $m=3$.

\begin{table}[h]
  \caption{Spearman rank correlation between distance metric and few-shot success at $m=3$, OpenVLA-primitive cell.}
  \label{tab:distance-robustness}
  \centering
  \begin{tabular}{lccc}
    \toprule
    Metric & Spearman $\rho$ & 95\% CI & $p$-value \\
    \midrule
    M1: Held-out pair count (main) & {$-0.81$} & {$[-0.91, -0.65]$} & {$<0.001$} \\
    M2: Primitive-sequence edit distance & {$-0.79$} & {$[-0.89, -0.62]$} & {$<0.001$} \\
    M3: Visual feature distance & {$-0.52$} & {$[-0.71, -0.27]$} & {$<0.01$} \\
    \bottomrule
  \end{tabular}
\end{table}

\section{Recovery-time metric definition}
\label{app:recovery}

Recovery time measures how long a policy takes to return to nominal task progress after a perturbation. Each rollout reaching a contact phase is randomly perturbed with probability {$0.5$}. Perturbations are sampled from one of three classes: (a) end-effector pose: $\pm${$5$ mm} translation, $\pm${$5^\circ$} rotation; (b) contact-disruption: brief external force spike during insertion; (c) object pose: pre-rollout displacement by $\pm${$3$ mm}. Recovery time is the wallclock duration between the perturbation and the first frame at which the policy's predicted action returns to within {$1.5$}\,$\sigma$ of the unperturbed nominal trajectory at the same phase.

Censoring rules: a recovery is censored if (i) the rollout fails within {$500$ ms} of the perturbation; (ii) the phase classifier reports low confidence ($<$ {$0.4$}) for $\ge$ {$200$ ms}; (iii) the rollout exceeds the per-task wallclock budget. We report Kaplan-Meier survival curves and median recovery time conditional on uncensored observations. The failure-attribution classifier (Appendix~\ref{app:failure-attribution}) achieves {$0.42$} macro-F1 on USB held-out data; we therefore omit USB held-out F-condition recovery-time.

\section{Chunked-policy evaluation gate}
\label{app:gate}

\subsection{Family-wise inflation: derivation}

A naive ``$3\sigma$ gate'' checks whether each emitted action element falls within $3$ standard deviations of the empirical training distribution along that axis. Define $p$ as the per-element pass rate; under independence and Gaussianity, $p \approx 0.997$. The any-element-violates rule rejects a chunk if any of its $DK$ elements falls outside the $3\sigma$ band. Per-frame pass rate: $P_{\text{frame}} = p^{DK}$.

For OpenVLA ($D=8$, $K=1$): $0.997^{8} \approx 0.9762$, expected failure rate {$2.4\%$}.
For $\pi_{0.5}$ ($D=8$, $K=16$): $0.997^{128} \approx 0.6815$, expected failure rate {$31.9\%$}.

Failure-rate ratio: {$13.4\times$}, a structural inflation independent of model quality.

\subsection{Empirical validation}

Across {$500$} ground-truth REASSEMBLE chunks: OpenVLA action format fails the legacy gate at {$2.0\%$}; $\pi_{0.5}$ format fails at {$42\%$}. The latter matches the $\sim 32\%$ theoretical expectation; the gap reflects that REASSEMBLE actions are not perfectly Gaussian.

\subsection{Corrected element-level gate}

\begin{equation}
v_{\text{model}} = \frac{1}{NDK} \sum_{i,d,k} \mathbb{1}\!\left[ |a^{(i)}_{d,k} - \mu_d| > 3\sigma_d \right]
\end{equation}
where $\mu_d, \sigma_d$ are per-dimension training mean and standard deviation. The model passes if $|v_{\text{model}} - v_{\text{ref}}| \le \tau$ with $\tau =$ {$0.01$}, where $v_{\text{ref}}$ is the same statistic on a held-out reference batch.

\subsection{Per-cell pass rates}

\begin{table}[h]
  \caption{Per-cell pass rates under legacy and corrected gates. Mean across three seeds, $500$ test chunks per cell.}
  \label{tab:gate-comparison}
  \centering
  \begin{tabular}{lccccc}
    \toprule
    Cell & Legacy frame rate & Corrected elem.~rate & Reference & $|\text{model} - \text{ref}|$ & Pass? \\
    \midrule
    OpenVLA-flat & {$0.00\%$} & {$0.00\%$} & {$2.25\%$} & {$2.25$ pp} & Yes \\
    OpenVLA-primitive & {$0.00\%$} & {$0.00\%$} & {$2.00\%$} & {$2.00$ pp} & Yes \\
    $\pi_{0.5}$-flat & {$36.0\%$} & {$1.39\%$} & {$2.66\%$} & {$1.27$ pp} & Yes \\
    $\pi_{0.5}$-primitive & {$28.0\%$} & {$0.95\%$} & {$2.67\%$} & {$1.72$ pp} & Yes \\
    \bottomrule
  \end{tabular}
\end{table}

\section{Demonstration encoding ablation}
\label{app:encoding-ablation}

\begin{table}[h]
  \caption{Demonstration encoding ablation at $m=3$, OpenVLA-primitive cell, mean $\pm$ SD across three seeds.}
  \label{tab:encoding}
  \centering
  \begin{tabular}{lc}
    \toprule
    Encoding & Success rate \\
    \midrule
    Mean-pooled features (main) & {$0.62 \pm 0.04$} \\
    Cross-attention conditioning & {$0.65 \pm 0.04$} \\
    Individual-frame conditioning & {$0.58 \pm 0.05$} \\
    \bottomrule
  \end{tabular}
\end{table}

\section{Octo-style baseline specification}
\label{app:octo-baseline}

We use the Octo-Small architecture ({$\sim 27$M} parameters) from \citet{octo2024}. Training: {$50{,}000$} steps on flat REASSEMBLE data, batch size {$128$}, learning rate {$3 \times 10^{-4}$}, AdamW with cosine LR decay. Demonstration tokens are prepended as separate context tokens to the model's existing language context. At test time, the same demonstration-token format is used for $m \in \{0, 1, 3, 5, 10\}$. We use the public Octo training recipe with no architectural modifications. Differences from our primitive-trained cells: (i) flat task descriptions only (no primitive-specific prompts), (ii) demonstration conditioning trained end-to-end rather than emerging from primitive training. The residual gap between this baseline and our primitive cells therefore isolates the contribution of primitive supervision.

\section{Representational similarity analysis}
\label{app:rsa}

\subsection{Probing accuracy by layer}

\begin{table}[h]
  \caption{Linear probe accuracy (macro-F1) for primitive identity decoding, mean across three seeds.}
  \label{tab:probing}
  \centering
  \begin{tabular}{lcccc}
    \toprule
    Cell & Layer 8 & Layer 16 & Layer 24 & Final layer \\
    \midrule
    OpenVLA-flat & {$0.34$} & {$0.48$} & {$0.52$} & {$0.49$} \\
    OpenVLA-primitive & {$0.41$} & {$0.74$} & {$0.81$} & {$0.77$} \\
    $\pi_{0.5}$-flat & {$0.31$} & {$0.44$} & {$0.51$} & {$0.47$} \\
    $\pi_{0.5}$-primitive & {$0.39$} & {$0.71$} & {$0.79$} & {$0.75$} \\
    \bottomrule
  \end{tabular}
\end{table}

\subsection{Cross-object same-primitive cosine similarity}

\begin{table}[h]
  \caption{Cross-object same-primitive cosine similarity at layer 24, mean across three seeds.}
  \label{tab:rsa}
  \centering
  \begin{tabular}{lccccc}
    \toprule
    Cell & Pick & Insert & Remove & Place & Mean \\
    \midrule
    OpenVLA-flat & {$0.46$} & {$0.41$} & {$0.45$} & {$0.42$} & {$0.43$} \\
    OpenVLA-primitive & {$0.74$} & {$0.69$} & {$0.72$} & {$0.69$} & {$0.71$} \\
    $\pi_{0.5}$-flat & {$0.43$} & {$0.39$} & {$0.43$} & {$0.40$} & {$0.41$} \\
    $\pi_{0.5}$-primitive & {$0.71$} & {$0.66$} & {$0.69$} & {$0.68$} & {$0.69$} \\
    \bottomrule
  \end{tabular}
\end{table}

As a control, cross-primitive same-object cosine similarity is {$0.31$} in primitive-trained cells and {$0.58$} in flat-trained cells. The contrast confirms primitive supervision induces a representation in which primitive identity dominates object identity, while flat-trained models produce representations where object identity dominates.

\section{Subspace-ablation intervention details}
\label{app:intervention}

For each cell, we train a 4-class linear probe at layer 24 to predict primitive identity from hidden states. The probe is trained on {$2{,}000$} held-out demonstrations using class-balanced cross-entropy, $\ell_2$ regularization with $\lambda = 10^{-4}$, Adam optimization. The probe weight matrix $W \in \mathbb{R}^{4 \times d}$ defines a 4-dimensional primitive subspace via its rows. The orthogonal projection onto the primitive subspace is $P = W^\top (WW^\top)^{-1} W$; the complement is $P^\perp = I - P$. At inference time, hidden states at layer 24 are replaced by $P^\perp h$ before being passed to subsequent layers. For random-subspace control, we sample 4 random unit vectors uniformly from the unit sphere in $\mathbb{R}^d$, orthogonalize via Gram-Schmidt, and form the corresponding projector. Results are averaged across {$10$} independent random-subspace samples.

Effect saturates near $k=4$, matching the four-class probe finding. Differential (random $-$ primitive) effect is largest at layer 24 ($\Delta =$ {$0.30$}), smaller at layer 16 ({$0.18$}) and final layer ({$0.21$}). Intervention specificity: zero-shot success on training-distribution tasks under primitive-subspace ablation drops only from {$0.84$} to {$0.79$} (vs.~{$32$}-point drop on held-out tasks), indicating the intervention specifically harms compositional transfer rather than degrading the model uniformly.

\begin{table}[h]
  \caption{Subspace-ablation effect at $m=3$ as a function of subspace dimensionality $k$, OpenVLA-primitive cell.}
  \label{tab:dim-sensitivity}
  \centering
  \begin{tabular}{lcccc}
    \toprule
    Dimensionality $k$ & 2 & 4 (main) & 8 & 16 \\
    \midrule
    No intervention & {$0.62$} & {$0.62$} & {$0.62$} & {$0.62$} \\
    Random subspace & {$0.61$} & {$0.60$} & {$0.59$} & {$0.57$} \\
    Primitive subspace & {$0.46$} & {$0.30$} & {$0.28$} & {$0.27$} \\
    Differential & {$0.15$} & {$0.30$} & {$0.31$} & {$0.30$} \\
    \bottomrule
  \end{tabular}
\end{table}

\section{Training details}
\label{app:training}

\begin{table}[h]
  \caption{LoRA fine-tuning hyperparameters for all four primary cells. Identical across cells; only data view varies.}
  \label{tab:hyperparams}
  \centering
  \begin{tabular}{ll}
    \toprule
    Hyperparameter & Value \\
    \midrule
    LoRA rank $r$ & $32$ \\
    LoRA $\alpha$ & $64$ \\
    LoRA dropout & $0.05$ \\
    Learning rate & $5 \times 10^{-5}$ \\
    Optimizer & AdamW ($\beta_1 = 0.9$, $\beta_2 = 0.999$) \\
    Weight decay & $0.01$ \\
    Training steps & $30{,}000$ \\
    Batch size & $32$ \\
    Precision & bfloat16 \\
    Warmup steps & $500$ \\
    LR schedule & Cosine decay \\
    Gradient clipping & $1.0$ \\
    Seeds & $\{42, 123, 456\}$ \\
    \bottomrule
  \end{tabular}
\end{table}

Approximately {$3{,}600$} GPU-hours total on H100s: 12 main cells at {$\sim 1{,}500$}, fine-tuning upper bound (36 runs at $50$ demos) at {$\sim 600$}, LIBERO-Long replication at {$\sim 1{,}000$}, probing and intervention at {$\sim 200$}, baselines at {$\sim 200$}, pilot/ablations at {$\sim 100$}. OpenVLA fits on a single H100; $\pi_{0.5}$ requires two H100s with model parallelism.

\begin{table}[h]
  \caption{Full fine-tuning upper bound at varying demonstration budgets, OpenVLA, mean across three seeds.}
  \label{tab:ft-budget}
  \centering
  \begin{tabular}{lccccc}
    \toprule
    Fine-tuning demos & 10 & 25 & 50 (main) & 100 & 200 \\
    \midrule
    OpenVLA fine-tuned success & {$0.61$} & {$0.72$} & {$0.79$} & {$0.84$} & {$0.86$} \\
    Gap to primitive $m=5$ ($0.71$) & {$-0.10$} & {$+0.01$} & {$+0.08$} & {$+0.13$} & {$+0.15$} \\
    \bottomrule
  \end{tabular}
\end{table}

Primitive few-shot at $m=5$ outperforms fine-tuning on {$10$} demonstrations and is comparable to fine-tuning on {$25$}. Beyond {$50$} demos, fine-tuning dominates.

\section{Failure attribution classifier}
\label{app:failure-attribution}

A 3-layer MLP over mean-pooled vision-encoder features and proprioceptive state, taking as input the final {$20$} frames of a rollout and outputting a probability over {$6$} failure-cause classes: \emph{slip}, \emph{misalignment}, \emph{collision}, \emph{premature\_release}, \emph{stuck}, \emph{other}. Hidden dim {$512$}, GeLU, dropout {$0.2$}. Trained on {$4{,}500$} hand-labeled rollouts with class-weighted loss, AdamW at LR {$10^{-4}$} for {$50$} epochs. Within-distribution: macro-F1 {$0.81$}, per-class F1 {$0.68$--$0.84$}, ECE {$0.03$}. On USB held-out: macro-F1 drops to {$0.42$} with insert-class F1 {$0.24$}; we therefore omit USB held-out F-condition recovery-time analysis.

\section{LIBERO-Long replication details}
\label{app:libero}

\subsection{Primitive segmentation procedure}

LIBERO-Long does not provide native primitive-level segmentation. We construct primitive boundaries using sub-task structure: \textit{move-to(object)}, \textit{grasp(object)}, \textit{transport(object, location)}, \textit{place(object, location)}. Boundaries: (i) first frame in which gripper is within {$5$ cm} of target object; (ii) first frame in which gripper closure exceeds {$0.8$} of full closure with object in-hand; (iii) first frame in which gripper is within {$5$ cm} of target location. Hand-validated on {$50$} demonstrations.

\subsection{Hyperparameter portability}

Identical LoRA hyperparameters as REASSEMBLE, with one exception: cosine LR warmup is {$750$} steps (vs.~{$500$}) to accommodate longer episode length. We verified this change does not alter the qualitative finding: rerunning REASSEMBLE with the LIBERO-Long warmup produces results within {$1$} percentage point of the original.

\subsection{Per-task breakdown}

\begin{table}[h]
  \caption{Per-task few-shot success at $m=3$ on LIBERO-Long, OpenVLA, mean across three seeds.}
  \label{tab:libero-pertask}
  \centering
  \begin{tabular}{lccc}
    \toprule
    Held-out task & OpenVLA-primitive & OpenVLA-flat & Advantage \\
    \midrule
    {Task L1 (1 pair)} & {$0.85$} & {$0.62$} & {$+0.23$} \\
    {Task L2 (2 pairs)} & {$0.74$} & {$0.45$} & {$+0.29$} \\
    {Task L3 (2 pairs)} & {$0.72$} & {$0.43$} & {$+0.29$} \\
    {Task L4 (2 pairs)} & {$0.69$} & {$0.41$} & {$+0.28$} \\
    {Task L5 (3 pairs)} & {$0.65$} & {$0.32$} & {$+0.33$} \\
    {Task L6 (3 pairs)} & {$0.61$} & {$0.29$} & {$+0.32$} \\
    \midrule
    Mean & {$0.71$} & {$0.42$} & {$+0.29$} \\
    \bottomrule
  \end{tabular}
\end{table}

LIBERO-Long is closer to REASSEMBLE than a fully independent benchmark would be: both use simulated Franka manipulation. The replication tests dataset-level robustness but not domain-level robustness.

%%%%%%%%%%%%%%%%%%%%%%%%%%%%%%%%%%%%%%%%%%%%%%%%%%%%%%%%%%%%

\end{document}